\documentclass[journal,twoside,web]{ieeecolor}
\usepackage{jsen}
\usepackage{cite}
\usepackage{amsmath,amssymb,amsfonts}
\usepackage{algorithmic}
\usepackage{graphicx}
\usepackage{textcomp}
\usepackage{wrapfig}
\usepackage{multirow}
\def\BibTeX{{\rm B\kern-.05em{\sc i\kern-.025em b}\kern-.08em
    T\kern-.1667em\lower.7ex\hbox{E}\kern-.125emX}}
\definecolor{abstractbg}{rgb}{0.89804,0.94510,0.83137}
\definecolor{gray}{rgb}{0.4, 0.4, 0.4}
\usepackage{eso-pic}
\usepackage{url}
\AddToShipoutPictureBG*{
  \AtPageUpperLeft{%
    \put(0,-40){\raisebox{18pt}{\makebox[\paperwidth]{\begin{minipage}{21cm}\centering
      \textcolor{gray}{This article has been accepted for publication in the IEEE Sensors Journal (JSEN). \\ DOI: 10.1109/JSEN.2024.3406948} 
    \end{minipage}}}}%
  }
  \AtPageLowerLeft{%
    \raisebox{20pt}{\makebox[\paperwidth]{\begin{minipage}{21cm}\centering
      \textcolor{gray}{ \copyright 2024 IEEE.  Personal use of this material is permitted.  Permission from IEEE must be obtained for all other uses, in any current or future media, including reprinting/republishing this material for advertising or promotional purposes, creating new collective works, for resale or redistribution to servers or lists, or reuse of any copyrighted component of this work in other works.
      }
    \end{minipage}}}%
  }
}

\begin{document}

\newcommand{\gi}[1]{\textcolor{gray}{\textit{#1}}} 

\title{Low Latency Visual Inertial Odometry with On-Sensor Accelerated Optical Flow for Resource-Constrained UAVs}

\author{Jonas Kühne, \IEEEmembership{Graduate Student Member, IEEE}, Michele Magno, \IEEEmembership{Senior Member, IEEE},\\ and Luca Benini, \IEEEmembership{Fellow, IEEE}
\thanks{This work was supported by the European Union’s ERA-NET CHIST-ERA 2018 Research and Innovation Programme APROVIS3D under Grant SNF-19 20CH21\_186991. \textit{(Corresponding author: Jonas Kühne.)}}
\thanks{Jonas Kühne is with both the Integrated Systems Laboratory and the Center for Project-Based Learning, ETH Zurich, 8092 Zurich, Switzerland (e-mail: kuehnej@ethz.ch). }
\thanks{Michele Magno is with the Center for Project-Based Learning, ETH Zurich, 8092 Zurich, Switzerland (e-mail: michele.magno@pbl.ee.ethz.ch).}
\thanks{Luca Benini is with the Integrated Systems Laboratory, ETH Zurich, 8092 Zurich, Switzerland, and also with the Department of Electrical, Electronic and Information Engineering, University of Bologna, 40136 Bologna, Italy (e-mail: luca.benini@unibo.it).}}

\IEEEtitleabstractindextext{%
\fcolorbox{abstractbg}{abstractbg}{%
\begin{minipage}{\textwidth}%
\begin{wrapfigure}[14]{r}{3in}%
\includegraphics[width=3in]{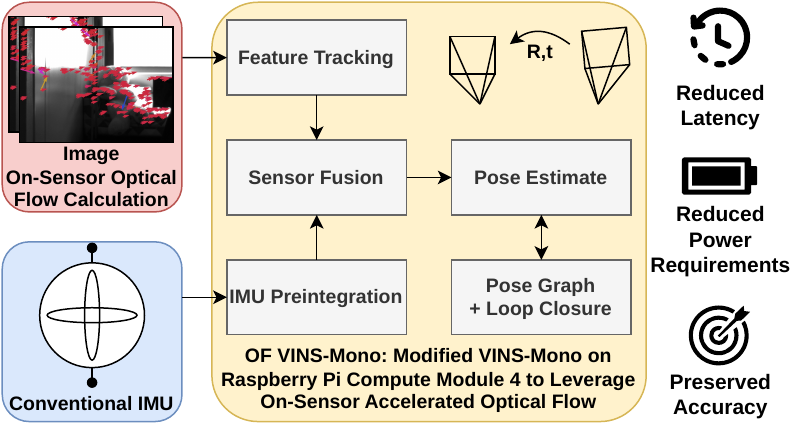}%
\end{wrapfigure}%
\begin{abstract}
\emph{Visual Inertial Odometry} (VIO) is the task of estimating the movement trajectory of an agent from an onboard camera stream fused with additional \emph{Inertial Measurement Unit} (IMU) measurements. A crucial subtask within VIO is the tracking of features, which can be achieved through \emph{Optical Flow} (OF). As the calculation of OF is a resource-demanding task in terms of computational load and memory footprint, which needs to be executed at low latency, especially in robotic applications, OF estimation is today performed on powerful CPUs or GPUs. This restricts its use in a broad spectrum of applications where the deployment of such powerful, power-hungry processors is unfeasible due to constraints related to cost, size, and power consumption. On-sensor hardware acceleration is a promising approach to enable low latency VIO even on resource-constrained devices such as nano drones. This paper assesses the speed-up in a VIO sensor system exploiting a compact OF sensor consisting of a global shutter camera and an \emph{Application Specific Integrated Circuit} (ASIC). By replacing the feature tracking logic of the VINS-Mono pipeline with data from this OF camera, we demonstrate a 49.4\% reduction in latency and a 53.7\% reduction of compute load of the VIO pipeline over the original VINS-Mono implementation, allowing VINS-Mono operation up to 50\,FPS instead of 20\,FPS on the quad-core ARM Cortex-A72 processor of a Raspberry Pi Compute Module 4.
\end{abstract}

\begin{IEEEkeywords}
Hardware acceleration, Optical flow, Visual odometry
\end{IEEEkeywords}
\end{minipage}}}

\maketitle

\section{Introduction}
Vision-based approaches to extract movement information are gaining popularity when complex navigation tasks need to be accomplished for the autonomous flight of unmanned aerial vehicles (UAVs) \cite{Lu2018,merzlyakov2021comparison}. Visual (inertial) odometry is commonly used, where features in the surroundings of the drone are being tracked \cite{delmerico2018benchmark,merzlyakov2021comparison}. These features or landmarks are then used to triangulate the position of the drone for every image frame to track the movement of the drone. Using the information of the visual odometry, the ego-motion of the drone can be estimated. This information is often augmented with additional sensor data for an absolute distance scale, more robust odometry estimation, and a higher position update rate \cite{qin2018vins}. For this purpose, inertial measurement units (IMUs) are a common choice \cite{delmerico2018benchmark,pu2023visual}. Fast visual odometry targeting a prediction rate of 100\,Hz and higher is challenging to implement as it has high requirements in terms of computational load, and memory usage \cite{mandal2019visual}. As a computationally affordable method, pre-integration of IMU measurements is often used to reach a prediction rate above 100\,Hz \cite{scaramuzza2019visual, forster2017manifold}. Moreover, even the fastest methods are today limited by the frame rates of the commercially available cameras, which are typically between 30 and 60 frames per second \cite{foehn2022agilicious}. Therefore today's cameras along with conventional VIO techniques are not fully adequate for high-speed robot applications such as fast UAV navigation \cite{lele2022fusing}. 

This work focuses on the low-power and low-latency tracking of image features for VIO using optical flow. Optical flow tracks the movement of features between consecutive frames in a stream \cite{ye2020unsupervised}. Compared to feature matching in arbitrary scenes this has the benefit that if the movement characteristic of the agent carrying the camera is known, then a maximum displacement can be defined to reduce the search range in feature matching \cite{zeng2021monocular}. 
The calculation of optical flow is a repetitive task that needs to be done several times per frame (usually for a fixed number of features) in a VIO pipeline. To achieve a low-power VIO system, this paper exploits a pre-commercial prototype of an optical sensor that can reach up to 300 frames per second \cite{kuehne2023fast}, designed by STMicroelectronics\footnote{www.st.com}. The sensor embeds hardware acceleration for optical flow computation, which allows a reduction of the computational load on the main processor, in combination with a Raspberry Pi Compute Module 4, and an IMU.

The computation of visual inertial odometry is an even more challenging but crucial task for nano-UAVs \cite{suleiman2019navion}. The growth in interest in such small drones has led to research on their use in narrow or inaccessible places (e.g. collapsed buildings, hydro-power dams, etc.) \cite{duisterhof2021sniffy,paliotta2021micro}. Navigation in tight, indoor spaces poses the challenge of unreliable or absent global positioning data, such as GPS or GLONASS \cite{duisterhof2021sniffy}. While small drones increase safety, due to their small scale and weight, they are also heavily restricted regarding the payload that can be carried \cite{Palossi2019}. This restricts the components that can be mounted on the UAV, most notably the battery, the computing unit, and its cooling, and the perception system, which forces the use of low-power resource-constrained processors that do not require active cooling \cite{Palossi2019}. In real-time applications, the use of low-power hardware leads to either reduced precision and robustness of the pose estimation or imposes overly strict constraints on other essential computing needs of the application such as the flight controller of the UAVs \cite{mandal2019visual}. Due to these restrictions, highly energy-efficient perception and computing solutions are required. One possible approach is to use accelerators in the form of application-specific integrated circuits (ASICs) for the processing of the perception data \cite{suleiman2019navion}. With our proposed VIO system which we termed OF VINS-Mono, we address these challenges. 

In particular, this paper addresses the challenge of resource-intensive Optical Flow (OF) calculations in Visual Inertial Odometry (VIO) for robotic applications \cite{li2023unmanned}. Specifically, it introduces a novel approach to enhance VIO performance by leveraging a novel compact optical flow sensor equipped with a global shutter camera and an integrated low-power ASIC, capable of achieving an impressive 300 frames per second directly on-sensor \cite{kuehne2023fast}. By integrating this sensor into the VIO pipeline, the paper demonstrates a substantial reduction in latency (49.4\%) and compute load (53.7\%) compared to the original VIO implementation (VINS-Mono). These improvements enable VINS-Mono to operate at up to 50 instead of 20 frames per second on a quad-core ARM Cortex-A72 processor, making it a promising solution for resource-constrained devices like nano drones, where traditional high-power processors are impractical due to cost, size, and power constraints.

In detail, the paper contributes the following results:
\begin{itemize}
    \item We propose a hardware-software codesign approach to reduce latency and power consumption and to reach higher tracking speeds in VIO systems using on-sensor OF estimation.
    \item We collect and open-source a VIO dataset containing the typical image and IMU data, plus the OF estimates of the camera \cite{kuehnej2024ofvio_dataset}. For the quantitative evaluation of the system, we record ground-truth poses with the VICON motion capture system.
    \item We assess the reduction of the processing load of an end-to-end low latency VIO pipeline on embedded processors when using on-sensor OF calculation. For our demonstration and quantification of the load-reduction, we use a Raspberry Pi Compute Module 4.
    \item We assess the potential speed-up that can be gained in a VIO setup with such an optical flow sensor by integrating it into the popular VINS-Mono pipeline \cite{qin2018vins} and demonstrate a latency reduction of 49.4\,\%.
    \item We compare the power requirements of the proposed VIO system that performs the OF estimation on the camera sensor versus the original VINS-Mono implementation where the feature tracking is performed on the CPU of the Raspberry Pi Compute Module 4.
\end{itemize}

\section{Related Work}

This section discusses related work in visual-inertial odometry systems. Since accelerated and low latency optical flow calculation is one of the critical enablers of this work, we discuss related work in the area of feature-based optical flow prediction by looking at feature detection, description, and matching algorithms. Furthermore, we present some of the recent VIO systems and we look at the efforts that have been made to build hardware accelerators for VIO systems.

\subsection{Feature Detection, Description, and Matching}

The detection of features on image frames is often performed with a layered approach, where first a fast algorithm is run to find candidate features. In the second step, a more sophisticated algorithm is used to determine if candidate points are suitable as feature points. These algorithms usually calculate the image gradient using either the Shi-Tomasi \cite{shi1994good} or the Harris \cite{harris1988combined} corner detector algorithm. If the image gradient is steep enough, a point is selected as a feature and the feature descriptor is calculated. As a last step, other nearby feature candidates are being suppressed to avoid describing the same feature multiple times. Modern algorithms \cite{rublee2011orbfeature} use feature detectors that are rotation invariant as well as scaling invariant within a certain range.

The feature descriptors, which are calculated on the de-rotated and re-scaled feature points, are usually either composed of a binary gradient representation as in SIFT \cite{lowe2004sift}, SURF \cite{bay2006surf}, or ORB \cite{rublee2011orbfeature}, or composed of the actual de-rotated and re-scaled image patch. 

The matching of features between frames is done by finding the most similar correspondence between two feature descriptors of two different frames, which can be consecutive (as in the calculation of optical flow) or not (as in place recognition or key-frame-based VIO). As a similarity metric for binary descriptors, the Hamming distance is used. For image patches, the difference between the template and the target patch is calculated \cite{zeng2021monocular}.

The optical flow sensor used in this work utilizes the FAST algorithm \cite{rosten2006machine} for the detection of features, which are described using the BRIEF descriptor \cite{calonder2010brief}. Matches between consecutive frames are found using the Hamming distance.

\subsection{Optical Flow Prediction and VIO Applications}

\begin{table}[t]
    \centering
    \caption{Related work in monocular visual (inertial) odometry.}
    \begin{tabular}{l|l|l|l}
        \hline
        \textbf{Algorithm} & \textbf{Published} & \textbf{VO/VIO} & \textbf{Method} \\
        \hline
        \textbf{PTAM \cite{klein2007parallel}} & 2007 & VO & Feature-based \\
        \textbf{DTAM \cite{newcombe2011dtam}} & 2011 & VO & Direct \\
        \textbf{SVO \cite{forster2014svo}} & 2014 & VO & Semi-direct \\
        \textbf{ORB-SLAM \cite{mur2015orb}} & 2015 & VIO & Semi-direct \\
        \textbf{DSO \cite{engel2017direct}} & 2017 & VO & Direct \\
        \textbf{VINS-Mono \cite{qin2018vins}} & 2018 & VIO & Feature-based \\
        \textbf{ORB-SLAM3 \cite{campos2021orb}} & 2021 & VIO & Semi-direct \\
        \hline
    \end{tabular}
    \label{tab:vio_algorithms}
\end{table}

Optical flow describes the displacement of a pixel or a feature from one image frame to the next. It can be therefore calculated by feature detection and matching as described previously. Alternatively, it can be directly calculated on a pixel patch by minimizing the photometric error, under the assumption that the brightness of the pixels in an image is nearly constant and the displacement between two images is small \cite{wang2018moving}. 

The same principles can be applied to VIO, i.e. the estimation of the movement of a camera between two image frames. Methods that reduce the photometric error are commonly termed \emph{direct methods}, whereas feature-based methods are sometimes called \emph{indirect methods}. Additionally, there is also a subset of algorithms that combine both approaches, by estimating the pose changes between keyframes in a feature-based manner and calculating pose changes to intermediate frames between two keyframes using a direct method, those algorithms are therefore called \emph{semi-direct} \cite{Lu2018}. An overview of influential monocular V(I)O implementations is given in Table \ref{tab:vio_algorithms}. The listed VIO pipelines are all designed to run on the CPU of the respective computing platform. While the field of VIO is active, most new pipelines are based on the listed algorithms, therefore we restrict the list to these highly cited publications.

This work extends the feature-based and highly modular VINS-Mono implementation by replacing the feature tracker with the optical flow data calculated on the VD56G3 sensor by STMicroelectronics.

\subsection{Low Power VIO Implementations}
Computer vision tasks have been a popular target for hardware acceleration, especially vision on drones has been tackled by various FPGA implementations \cite{lentaris2015hw,tertei2016fpga,zhang2017visual}. Although the computing platforms have become much more powerful ever since the interest in smaller and smaller drones has led to continuous research in this area \cite{wan2021survey}. 

Especially in the area of VIO, different approaches have been taken to optimize the computation and power overhead through hardware acceleration using FPGAs. \cite{lentaris2015hw} presents multiple visual odometry pipelines where different sub-processes of the feature detection and matching pipeline are outsourced to an FPGA. On the contrary, \cite{tertei2016fpga} presents an implementation where the EKF-based pose estimation is accelerated and outsourced to an FPGA. As an additional example in \cite{zhang2017visual} the authors take a more holistic approach in accelerating a stereo vision VIO pipeline which uses iterative pose graph optimization.

In \cite{bahnam2021stereo} the authors take a different approach by heavily optimizing the popular S-MSCKF algorithm \cite{sun2018robust} towards the used computing platform, a Raspberry Pi Zero. By reducing the number of tracked features and simplifying the stereo matching to only the x-direction the authors are able to half the processing time.

In our work, we use an accelerator for the optical flow calculation implemented on an ASIC, while running the remaining parts of the VIO pipeline on the main processor of a Raspberry Pi Compute Module 4.

\begin{figure*}[t]
    \centering
    \includegraphics[width=\linewidth]{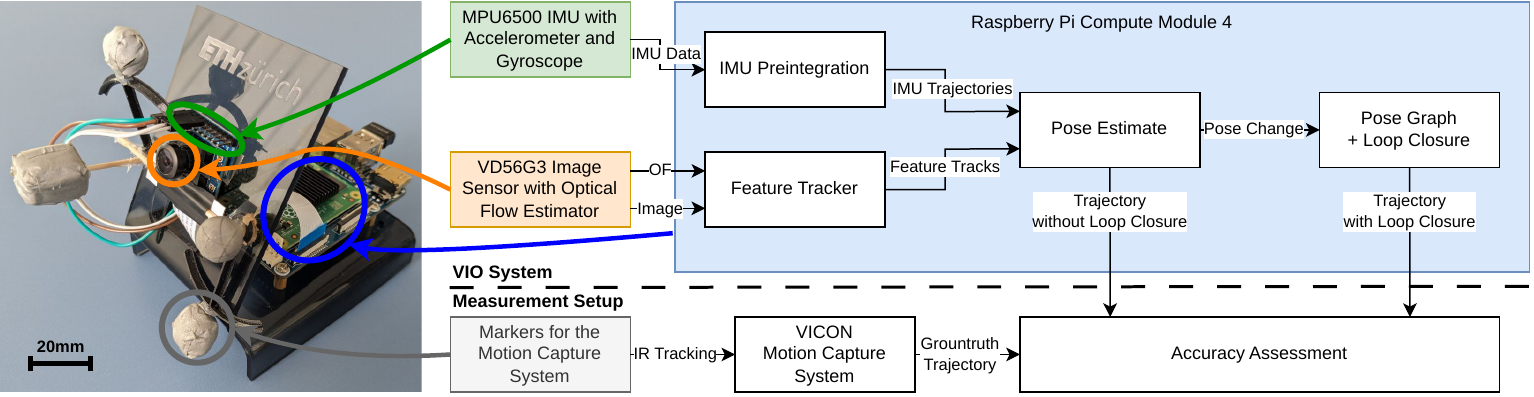}
    \caption{The hardware system consists of the VD56G3 sensor, an MPU6500 IMU, and a Raspberry Pi Compute Module 4. Markers were placed on the sensor to record ground truth poses with a VICON motion capture system. The main software components of the OF VINS-Mono pipeline consist of the IMU pre-integrator module, the feature tracker, which concatenates optical flow vectors to feature tracks, the pose estimation module which fuses the sensor information of the optical flow sensor together with the IMU data to estimate pose changes, and the pose graph module, which handles loop closures and optimizes the obtained pose graph.}
    \label{fig:system_overview}
\end{figure*}

\section{The VIO System}
\label{sec:vio_system}
In this work, we propose a hardware-software codesign approach to augment VIO systems with on-sensor OF estimation. Furthermore, we present a VIO system implementing our approach by extending the popular VINS-Mono \cite{qin2018vins} pipeline. This new system reduces the power requirements while reaching a lower latency compared to the baseline VINS-Mono implementation through the use of a fast and energy-efficient optical flow sensor. Outsourcing the calculation of the optical flow to the camera sensor furthermore frees up computational resources in the host allowing either for the use of a less powerful computing platform or enabling more computationally demanding subsequent planning and control tasks.

In the following subsections, we describe the hardware-software codesign approach as well as the selected hardware components, including the relevant configuration parameters. Furthermore, we describe how we interface with the existing VINS-Mono pipeline at the software level.

\subsection{System Overview}
\label{sec:system_overview}
The proposed system depicted in Fig. \ref{fig:system_overview} consists of the following hardware components. The VD56G3 optical flow sensor by STMicroelectronics is used for the acquisition of monochrome images and optical flow predictions. The VD56G3 sensor contains a monochrome pixel array with an integrated ASIC for the computation of optical flow vectors \cite{kuehne2023fast}. As a complementary Inertial Measurement Unit (IMU) an MPU6500 by TDK is being used. A Raspberry Pi Compute Module 4 with 2\,GB of RAM has been chosen as the host computer and processing platform for the remaining VIO pipeline steps. However, the proposed approach can be scaled up or down for use in combination with other embedded processors. When battery-powered, the system can be operated as a self-contained module.

\subsection{The VD56G3 Image Sensor}
The VD56G3 is a 1.5-megapixel global shutter image sensor with integrated optical flow motion vector computation in hardware. The image sensor has an 1124-pixel by 1364-pixel resolution and produces monochrome images at up to 88 frames per second at the full resolution. At lower resolutions, the image sensor can reach frame rates as high as 300 frames per second. To get to lower-resolution images, the original image can be either cropped, sub-sampled, or a combination of both. The image sensor supports 2x and 4x sub-sampling and binning, as well as almost arbitrary cropping, through an area of interest setting. The optical flow unit can operate on input images up to VGA size (640 by 480 pixels), therefore the input image is automatically down-sampled if it is bigger than the supported size before computing the OF vectors. The optical flow unit could theoretically predict up to 2048 motion vectors, which can be achieved by setting the desired number of BRIEF descriptors up to 2048. After the matching with the BRIEF descriptors of the previous frame, the number of effectively found matches and therefore optical flow vectors is typically significantly lower than the number of the BRIEF descriptors as not all the features can be matched. This can be due to certain features moving out of the visible frame, or becoming occluded, additionally, the feature appearance could also change significantly due to lighting changes, such that it can no longer be matched. 

The user can selectively only transmit image data, optical flow data, or both to the host computer. Multiple regions of interest (ROI) can be defined, and with those settings distinct image regions can be selected for the creation of the image that will be transmitted, the optical flow calculation, and the auto exposure controller. In our experiments, we kept all three ROIs identical. It is worth mentioning that only the OF vectors, but not the previously calculated BRIEF descriptors are being exposed across the chip's interface. Depending on the image resolution and the number of optical flow vectors that need to be processed, different frame rates can be reached. The OF calculation on the sensor is performed in parallel to the image processing and transmission. Therefore, the OF calculation adds no additional latency other than the time required to transmit the OF vectors to the host computer. The maximum frame rates for some typical resolutions are given in Table \ref{tab:framerates}. Since VGA is the largest resolution at which the OF unit can be operated without subsampling the incoming images, we used this resolution for our experiments.

\begin{table}[t]
    \centering
    \caption{Achievable frame rates for a given frame height and a given number of optical flow vectors.}
    \begin{tabular}{c|c|c|c}
        \hline
        \textbf{Format} & \textbf{Frame Height [pixel]} & \textbf{\# OF Vectors} & \textbf{Frame Rate [1/s]} \\
        \hline
        QVGA & 240 & 1024 & 338 \\
        QVGA & 240 & 2048 & 288 \\
        VGA & 480 & 0 & 229 \\
        \textbf{VGA} & \textbf{480} & \textbf{1024} & \textbf{205} \\
        VGA & 480 & 2048 & 186 \\
        FULL & 1364 & 0 & 88 \\
        FULL & 1364 & 1024 & 84 \\
        FULL & 1364 & 2048 & 80 \\
        \hline
    \end{tabular}
    \label{tab:framerates}
\end{table}

\subsection{OF VINS-Mono}
We selected VINS-Mono \cite{qin2018vins} as a baseline for this work due to its modular nature. The clear separation of feature tracker, VINS estimator (i.e. local visual-inertial odometry with relocalization), and global pose graph logic, allows for easy customization and hardware acceleration of subroutines.

Another useful feature of VINS-Mono is the fact, that the VINS estimator and the pose graph logic do not rely on the feature descriptors obtained in the feature tracker. These modules instead recalculate feature descriptors when needed for relocalization. As previously mentioned, the VD56G3 does not expose the feature descriptors obtained during the optical flow estimation, therefore necessitating a recalculation of feature descriptors for relocalization and loop-closure as is done in VINS-Mono.

To adapt VINS-Mono to work in conjunction with the optical flow sensor, the original feature tracker was removed completely. As optical flow data is being generated directly on the camera sensor, logic was added as C++ code for the pre-processing of the optical flow vectors, i.e. the concatenation of optical flow over several frames to feature tracks, as well as for the undistortion and projection to a unit sphere as required by VINS-Mono \cite{qin2018vins}. This pre-processing logic is fully compatible with the VINS-Mono interface, and the data can be fed directly to the VIO estimator of VINS-Mono. VINS-Mono rejects outliers using the fundamental matrix model with RANSAC \cite{qin2018vins}. In essence, the fundamental matrix model describes a projection of feature points in one image frame to a different image frame of the same scene. Using RANSAC, this fundamental matrix is determined for a pair of images using feature point correspondences, the features that do not agree with the fundamental matrix transformation are then removed. This method helps to suppress features from non-static objects, which is necessary in a purely monocular case but can be omitted in VIO as the IMU provides an additional movement hypothesis, with which the outliers can be filtered out in a later stage. Hence, in our adaptation we do not implement any specific outlier detection, instead, we select the 150 longest feature tracks and add new optical flow vectors with a low Hamming distance score, whenever older tracks are lost.

\begin{table*}[t]
    \centering
    \caption{3D-RMSE and standard deviation of the pose estimation error in meters when running both VIO systems (OF VINS-Mono and Original VINS-Mono) on the recordings of the dataset, both with and without loop closure (LC). The trajectories have been aligned with the ground truth using both the sim(3) and pose yaw transformation of \cite{zhang2018tutorial}.}
    \begin{tabular}{l|c|c|l|l|l|l|l}
    \hline
    \textbf{Algorithm}  & \textbf{Alignment} & \textbf{LC} & \multicolumn{5}{c}{\textbf{RMSE and \gi{Standard Deviation of the Error} per Parameter Set [m] $\downarrow$}} \\
    \hline
     & & & 20 FPS & \emph{10 FPS}  & 20 FPS & 20 FPS & 20 FPS \\
     & & & 200 BRIEF & 200 BRIEF & \emph{150 BRIEF} & \emph{300 BRIEF} & 200 BRIEF \\
     & & & 4 Spatial$^\ast$ & 4 Spatial$^\ast$ & 4 Spatial$^\ast$ & 4 Spatial$^\ast$ & \emph{2 Spatial$^\ast$} \\
    \hline
    OF       & sim(3)   & no  & 0.540 \gi{0.23} (-29.8\%)          & 1.468 \gi{0.55} (+1.7\%) & \textbf{0.309} \gi{0.13} \textbf{(-52.2\%)} & 0.657 \gi{0.31} (+26.3\%) & 0.751 \gi{0.28} (+14.1\%) \\
    Original & sim(3)   & no  & 0.769 \gi{0.30}                    & 1.443 \gi{0.60}          & 0.647 \gi{0.29}                    & 0.520 \gi{0.21}           & 0.658 \gi{0.24} \\
    OF       & sim(3)   & yes & \textbf{0.511} \gi{0.25} (-34.3\%) & \textbf{1.365} \gi{0.60} (-0.2\%) & 0.423 \gi{0.13} (-32.0\%) & 0.720 \gi{0.40} (+58.2\%) & 1.014 \gi{0.42} (+75.4\%) \\
    Original & sim(3)   & yes & 0.778 \gi{0.31}                    & 1.368 \gi{0.64}                   & 0.622 \gi{0.29}           & \textbf{0.455} \gi{0.20}  & \textbf{0.578} \gi{0.29} \\
    \hline
    OF       & pose yaw & no  & 0.592 \gi{0.27} (-26.7\%) & 2.601 \gi{1.23} (+53.8\%) & 0.342 \gi{0.17} (-49.6\%) & 0.743 \gi{0.33} (+32.4\%) & 0.947 \gi{0.41} (+19.4\%) \\
    Original & pose yaw & no  & 0.808 \gi{0.32}           & 1.691 \gi{0.93}           & 0.678 \gi{0.31}           & 0.561 \gi{0.21}           & 0.793 \gi{0.23}  \\
    OF       & pose yaw & yes & 0.562 \gi{0.26} (-30.2\%) & 1.975 \gi{0.98} (+22.0\%) & 0.493 \gi{0.19} (-22.2\%) & 0.958 \gi{0.50} (+106\%)  & 1.400 \gi{0.66} (+122\%) \\
    Original & pose yaw & yes & 0.805 \gi{0.35}           & 1.619 \gi{0.97}           & 0.634 \gi{0.29}           & 0.465 \gi{0.20}           & 0.632 \gi{0.33} \\
    \hline
    \multicolumn{8}{l}{$^\ast$n Spatial denotes the maximum spatial density of BRIEF descriptors, i.e. a maximum of n BRIEF descriptors per 16 by 16-pixel patch.} \\
    \end{tabular}
    \label{tab:rmse}
\end{table*}

\section{Methodology}
We tested and benchmarked our proposed system against the original VINS-Mono pipeline as proposed in \cite{qin2018vins}. To obtain comparable results we ran multiple experiments on the hardware platform presented in section \ref{sec:vio_system}. 

\subsection{Tracking Accuracy}
\label{sec:method_accuracy}
To evaluate the tracking accuracy of both the newly proposed OF VINS-Mono and the original VINS-Mono we recorded a VIO dataset containing monochrome image data, IMU data, and the optical flow data predicted by the VD56G3 \cite{kuehnej2024ofvio_dataset}. To obtain ground-truth pose measurements we used a VICON motion capture system running at 100\,Hz. The dataset was created using five different parameter sets on the VD56G3 sensor, of which two have an influence on the frame rate and hence impact the tracking accuracy of both OF and original VINS-Mono, and three additional parameter sets that impact the quality of the optical flow predictions and hence only influence the tracking accuracy of the proposed OF VINS-Mono pipeline. The varied parameters are:

\begin{enumerate}
    \item The frame rate (FPS) at which both the image frames are being recorded and the optical flow vectors are being predicted.
    \item The number of desired BRIEF descriptors, the VD56G3 has an internal controller that tunes the cornerness threshold of the optical flow pipeline to reach the desired number of descriptors.
    \item The number of BRIEF descriptors per 16 by 16-pixel patch, which can be used to steer the spatial density of the optical flow predictions.
\end{enumerate}

The baseline parameter set uses 20 FPS, a target of 200 BRIEF descriptors, and a maximum of 4 BRIEF descriptors per 16 by 16-pixel patch (denoted 4 Spatial), all five parameter sets are given in Table \ref{tab:rmse}.

\begin{figure}[t]
    \centering
    \includegraphics[width=\linewidth]{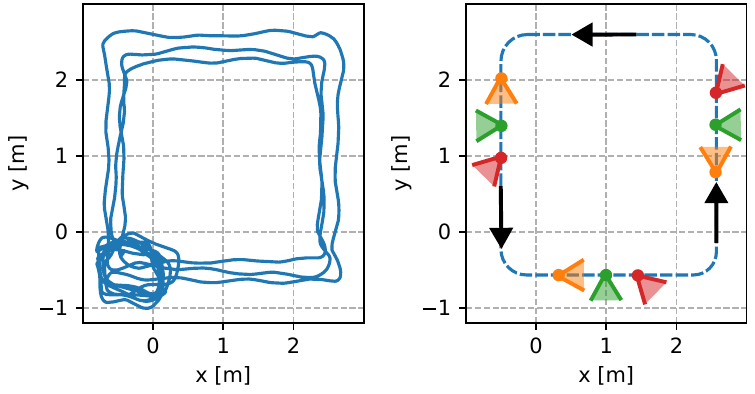}
    \caption{\textbf{Left:} Sample trajectory (in blue) of the movement in the 4-meter by 4-meter room captured by the VICON system. The smaller loops in the bottom-left corner of the plot were performed to align the ground-truth recording with the predictions of both VINS-Mono systems. \\ \textbf{Right:} The three different camera orientations are indicated: in the movement direction (orange), perpendicular to the movement direction (green), and at 45 degrees to the movement direction (red).}
    \label{fig:sample_trajectory}
\end{figure}

For each of the five parameter sets defined in Table \ref{tab:rmse}, we recorded four sequences that contain movement within a square of 4 meters by 4 meters (limited by the capturing range of the used VICON system). The first sequence contains a mix of movements with the camera pointing either in the movement direction or perpendicular to the movement direction, including various turns. This first sequence represents the general movement as encountered in real-world drone flight. For the second to fourth movement sequences the camera is pointing in a pre-defined direction per sequence as shown in Fig. \ref{fig:sample_trajectory}: 
\begin{itemize}
    \item for the second sequence the camera is pointing in the movement direction,
    \item for the third sequence the camera is oriented perpendicular (horizontally) to the movement direction, and
    \item for the fourth sequence the camera is oriented at an angle of 45 degrees (horizontally) to the movement direction.
\end{itemize}

The recordings were used as input to both the original VINS-Mono and the newly proposed OF VINS-Mono pipeline, to obtain the predicted poses. As the VD56G3 sensor does not expose the feature descriptors, it is not possible to run loop-closure on the feature descriptors generated by the optical flow sensor, therefore the loop-closure logic uses the monochrome images recorded by the VD56G3 sensor to newly generate feature descriptors. To obtain a complete picture we ran our tests both with and without loop-closure. The resulting trajectories of the first sequence were aligned with the ground truth poses obtained with the VICON system based on the timestamps of the poses, both using a similarity transformation (sim(3)) and a pose yaw transformation of \cite{zhang2018tutorial}, before calculating the three-dimensional root mean square errors with respect to the ground truth. The resulting errors are given in Table \ref{tab:rmse}. These error terms are used as an indicator of the trajectory similarity produced by the two VINS-Mono implementations versus the VICON ground truth. To improve the comparability of our work with previous works that only report one of the metrics, we decided to include both, sim(3) as well as pose yaw transformation.

In an additional ablation study, we used the movement sequences two to four to analyze the drift of both OF VINS-Mono and original VINS-Mono under the various camera orientations. To do so the predicted trajectories are aligned for the first 15 seconds of the recording using the pose yaw transformation. The trajectories then diverge for the remaining length of the recording, hence loop-closure is deactivated. To quantify the amount of drift for both models and the various parameter sets, we then calculate rotational and end-point errors for various trajectory lengths.

\subsection{Latency}
As one of our main goals is to reduce the latency of the pose estimation in our proposed OF VIO system, we compared the latency of OF VINS-Mono with the original VINS-Mono implementation by running both systems on identical hardware.

We analyzed the changes in latency of the software parts on a Raspberry Pi Compute Module 4. As both VINS pipelines are implemented in ROS\footnote{www.ros.org}, we measure the time needed from the reception of new data until the publishing of the processed data is done (i.e. from reception of data until the processed data is sent out). The time is being measured using the \emph{rostime} primitives. Our latency analysis also considers changes in latency in the subsequent processing step after feature tracking namely the pose estimation, to investigate if the difference in feature tracking algorithms from the original to the OF VINS-Mono algorithm also has an impact on later stages in the VIO pipeline.

For the sake of completeness, we analyze the time needed for the transmission of the image plus the additional overhead caused by the transmission of the OF vectors. While the transmission of the image requires a constant amount of time for a fixed frame rate, the transmission of the OF vectors can vary depending on the number of found feature matches. The VD56G3 camera allows setting a maximum number of BRIEF descriptors, therefore we can indicate a pessimistic upper bound on the time requirement.

\subsection{Power Consumption}
To verify that hardware acceleration is sensible from a power perspective, we compare the power consumption of the original VINS-Mono system versus the proposed OF VINS-Mono system. To do so, the power consumption of the entire system is measured, whereas, for the original VINS-Mono system, the optical flow unit of the VD56G3 sensor is disabled. For OF VINS-Mono the optical flow unit of the VD56G3 sensor is activated, and the more lightweight OF VINS-Mono implementation is run on the Raspberry Pi CM4. It is worth noting that the Raspberry Pi runs a full Raspbian OS installation, therefore the difference in power consumption from the original VINS-Mono to the OF VINS-Mono implementation is a better performance metric, than the absolute power draw. To improve the comparability of the results, we made sure that the idle power consumption was similar for both measurement runs.

The power draw of the system has been measured with an N6705C power analyzer by Keysight. The Raspberry Pi CM4 has been powered via one of the 5-volt supply GPIOs and both the VD56G3 OF camera and the MPU6500 IMU have been supplied by the 3.3-volt rail of the Raspberry Pi CM4. To account for variances in the power draw caused by the Raspbian OS, we measured the idle power consumption of both VIO pipelines. Additionally, we also measured the power draw when only the camera capture was active. In the case of the original VINS-Mono pipeline the OF unit on the VD56G3 is disabled, whereas for OF VINS-Mono the OF unit is active.

\section{Results}
This section presents the results that have been obtained by measuring tracking accuracy, latency, and system power draw of the proposed OF VINS-Mono system versus the original VINS-Mono pipeline. For the evaluation of the tracking accuracy the newly collected dataset has been used.

\subsection{Accuracy of VINS-Mono}
Running both VIO pipelines on the first movement sequence for each of the five different parameter sets and analyzing them using the methods described in Section \ref{sec:method_accuracy} yielded the results shown in Table \ref{tab:rmse}. Reducing the frame rate, which affects both the original and the OF VINS-Mono pipeline, leads to a consistent degradation of tracking accuracy for both systems. For the remaining experiments with a fixed frame rate of 20\,FPS only the configuration of the optical flow unit of the VD56G3 is changed, hence only affecting the performance of the OF VINS-Mono implementation. From Table \ref{tab:rmse}, a variance in RMSE for the original VINS-Mono system between 0.520\,m and 0.769\,m can be observed across the experiments without loop closure, indicating that it is challenging to reproduce a VIO recording. Therefore, we indicate also the relative reduction or increase in RMSE of the OF VINS-Mono system over the original implementation. From the relative performance change, it can be seen that both configurations with a low BRIEF target (BRIEF 150 and 200) and a spatial limit of 4 BRIEF descriptors per 16 by 16-pixel patch lead to a decreased RMSE of 52.2\% and 29.8\% respectively. More specifically, the tracking performance of OF VINS-Mono decreases when more BRIEF descriptors are being generated, both in absolute terms and relatively in comparison to the original VINS-Mono implementation. Since the BRIEF target implicitly tunes the cornerness threshold in the feature detection, this indicates that the BRIEF target acts as a strong filter for outliers, whereas when using a high BRIEF target, only selecting features based on the feature track length is not an ideal outlier removal scheme. Both the performance of the baseline parameter set, as well as the best-performing parameter set (BIREF 150), indicate that OF VINS-Mono can compete with the original VINS-Mono implementation and surpass it when appropriately configured.

\begin{table}[t]
    \centering
    \caption{3D-RMSE and standard deviation of the pose estimation error in meters when running both VIO systems (OF VINS-Mono and Original VINS-Mono) with specific camera orientations. The trajectories have been aligned with the ground truth for the first 15 seconds of the trajectory using the pose yaw transformation of \cite{zhang2018tutorial}.}
    \begin{tabular}{l|l|l|l}
    \hline
    \textbf{Camera Orientation} & \multicolumn{3}{c}{\textbf{RMSE and \gi{Standard Deviation} [m] $\downarrow$}} \\
    \hline
     & 150 BRIEF & 200 BRIEF & 300 BRIEF \\
    \hline
    Front Facing  &  1.031 \gi{0.57} & 0.729 \gi{0.41} & 0.472 \gi{0.27} \\
    Perpendicular &  0.432 \gi{0.13} & 1.678 \gi{1.11} & 31.441 \gi{14.07} \\
    45 Degrees    &  3.819 \gi{2.42} & 1.238 \gi{0.76} & 2.402 \gi{1.21} \\
    \hline
    \end{tabular}
    \label{tab:rmse_ablation}
\end{table}

\begin{figure}[t]
    \centering
    \includegraphics[width=\linewidth]{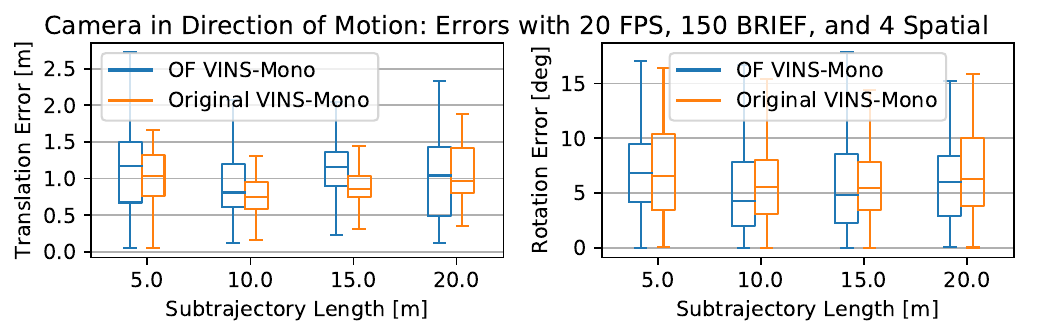}
    \includegraphics[width=\linewidth]{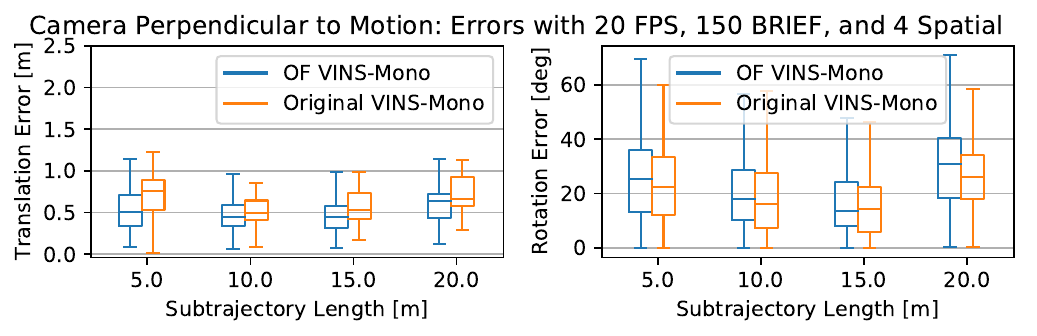}
    \caption{The plots show the translational and rotational errors of randomly sampled sub-trajectories of the indicated lengths for the BRIEF 150 parameter set. The top plots correspond to the camera pointing in the direction of movement, whereas in the bottom plot, the camera is oriented perpendicularly to the movement direction. Although the rotational drift in the top plot is much smaller, we observed that it is systematic for both the original VINS-Mono and OF VINS-Mono, leading to an accumulation of the error, which is also reflected in the larger translation error compared to the bottom plot.}
    \label{fig:vins_errors}
\end{figure}

The ablation study shown in Table \ref{tab:rmse_ablation}, in which we analyzed single movement patterns, confirms the overall finding: the baseline parameter set and BRIEF 150 perform best. However, it can be additionally observed, that certain settings perform well in specific movement patterns. With the parameter set BRIEF 300 for example, OF VINS-Mono (3D-RMSE of 0.472\,m) exceeds the accuracy of the original VINS-Mono significantly (3D-RMSE of 0.978\,m) when the camera is pointed in the movement direction. On the contrary OF VINS-Mono with BRIEF 300 diverges and fails when the camera is pointed perpendicular to the movement direction. Similarly, we can observe a large RMSE for BRIEF 150 when the camera is oriented at 45 degrees with respect to the direction of movement. In this setting, OF VINS-Mono tends to overestimate the rotation in corners with few features, which leads to a strong drift of the predicted trajectory. In scenarios where the camera is oriented perpendicular to the movement direction, which results in large relative movement of the scene, OF VINS-Mono can outperform the original VINS-Mono when adequately configured, as indicated in Fig. \ref{fig:vins_errors}.

\subsection{Speed-Up and Latency Reduction}

The original vs. OF VINS-Mono implementations differ in two main aspects that influence the latency: 1. the added transmission latency of the OF vectors in OF VINS-Mono and 2. the reduced latency in the feature tracking of OF VINS-Mono due to the outsourcing to the sensor. More specifically, in OF VINS-Mono, the computation of the OF happens in parallel to the transmission of the image data, so no extra latency is caused in comparison to the original VINS-Mono implementation. However, the transmission of the OF vectors requires additional time. Subsequently, we detail the latency aspects both of the hardware and software components.

For our experiments, we used a MIPI CSI-2 transmission rate of 804\,Mbps as recommended by STMicroelectronics for the use with a Raspberry Pi. Image data is encoded with 8\,bits\,per\,pixel. The resulting image transmission latency for both VINS-Mono systems is therefore 3.06\,ms (i.e. VGA image with the specified encoding and data rate).

The added hardware latency for the transmission of the OF vectors, which are 8 bytes long, amounts to the following worst-case latencies: 0.016\,ms for a target of 150 BRIEF descriptors (maximum of 200 descriptors), 0.024\,ms for a target of 200 BRIEF descriptors (maximum of 300 descriptors), and 0.04\,ms for a target of 300 BRIEF descriptors (maximum of 500 descriptors).

To analyze the software timing characteristics of the two different VINS-Mono implementations, we ran both on the Raspberry Pi Compute Module 4. We indicate the time required for the feature update and the VINS-Estimation (i.e. pose estimation) separately. For the original implementation, the feature update includes feature detection, description, and outlier rejection. The OF VINS-Mono implementation only includes the undistortion and feature track update.

The original VINS-Mono system defines a prediction rate for pose estimation, which can be different from the actual frame rate of the (optical flow) camera. The feature tracking node performs certain operations at the same rate as the camera's frame rate (i.e. updating the feature tracks) and sends the processed information to the VINS-Estimation only with the desired prediction rate. Therefore we separately indicate those timings in rows (A) and (B) of Table \ref{tab:timing} respectively. As the selection of the features might also influence the processing time of the subsequent pose estimation, we indicate the timing characteristics of the pose estimation (VINS-Estimator) in row (C) of Table \ref{tab:timing} for the two implementations.

\begin{table}[t]
    \centering
    \caption{Timing Characteristics of (A) the feature update without sending the data to the pose estimator, (B) the feature update including sending the data to the pose estimator, and (C) the pose estimation (VINS-Estimator).}
    \begin{tabular}{ll|c|c|c|c}
        \hline
        & \textbf{Method} & \textbf{Mean [ms] $\downarrow$} & \textbf{SD [ms]} & \textbf{Min [ms]} & \textbf{Max [ms]}  \\
        \hline
        \multirow{2}{*}{(A)} & \textbf{OF} & \textbf{0.3390} & 0.1678 & 0.0280 & 3.7270 \\
        & Original & 14.2538 & 2.2102 & 8.9040 & 37.9660 \\
        \hline
        \multirow{2}{*}{(B)} & \textbf{OF} & \textbf{0.4932} & 0.2243 & 0.0750 & 4.5060 \\
        & Original & 60.7917 & 13.7801 & 14.6850 & 104.7460 \\
        \hline
        \multirow{2}{*}{(C)} & \textbf{OF} & \textbf{74.3676} & 29.7808 & 1.0530 & 901.7000 \\
        & Original & 87.3183 & 69.4731 & 0.8760 & 2124.7760 \\
        \hline
    \end{tabular}
    \label{tab:timing}
\end{table}

\begin{figure}[t]
    \centering
    \includegraphics[width=\linewidth]{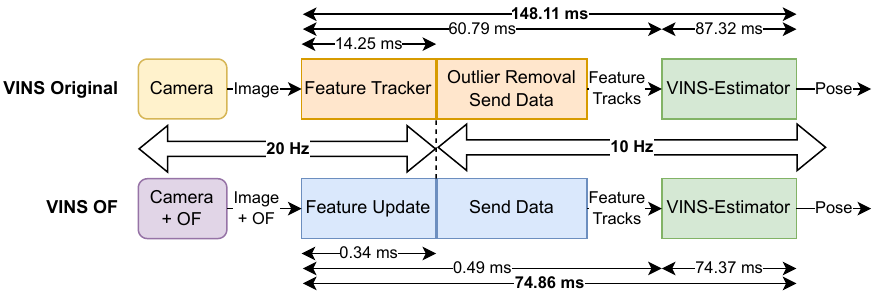}
    \caption{Latency breakdown for the calculation of one odometry estimation for both implementations. The IMU pre-integration is omitted in the diagram for better readability.}
    \label{fig:time_breakdown}
\end{figure}

In our experiments, we were able to reduce the mean processing time from image frame to odometry estimation from 148.11\,ms to 74.86\,ms, reducing the average latency of the odometry pipeline by 49.4\% as illustrated in Fig. \ref{fig:time_breakdown}. As in the baseline parameter set both the camera and feature update are operated at 20\,FPS, whereas the pose estimation (VINS-Estimator) is per default run at 10\,Hz we can reduce the average processing time per odometry estimation, i.e. two feature updates plus one VINS-Estimation, from 162.36\,ms to 75.20\,ms, resulting in a reduction of 53.7\%. The reduced time for the pose estimation is caused partially by the better quality of the features of OF VINS-Mono and partially by the reduced overall load on the processor by outsourcing the optical flow computation to the ASIC on the camera.

As the feature tracking logic of OF VINS-Mono is only adding limited overhead both in hardware and software the high frame rates of the VD56G3 camera can be used to track fast motions as long as the VINS-Estimation is kept at a moderate rate.

\subsection{System Power Draw}
\begin{figure}[t]
    \centering
    \includegraphics[width=\linewidth]{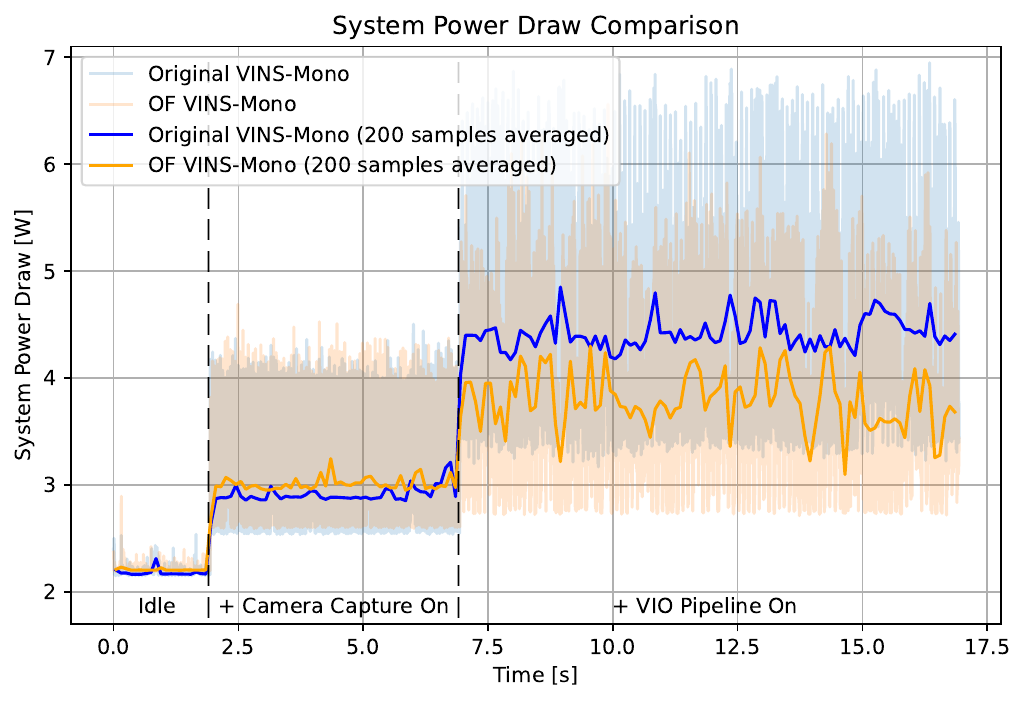}
    \caption{A comparison of the system power draw of the original VINS-Mono pipeline versus OF VINS-Mono. Three phases are being shown, the idle consumption of the Raspberry Pi CM4, the power consumption when only the camera capture is enabled and the power draw when either VIO pipeline is being fully operational. The plot shows both the averaged (opaque lines) and non-averaged (transparent lines) system power draw.}
    \label{fig:power_draw}
\end{figure}

\begin{table}[t]
    \centering
    \caption{Average Power Consumption of the two VIO Systems.}
    \begin{tabular}{l|c|c|c}
        \hline
        \textbf{System} & \textbf{Idle [W] $\downarrow$} & \textbf{+ Camera [W] $\downarrow$} & \textbf{+ VIO [W] $\downarrow$} \\
        \hline
        OF & 2.21 & 3.01 & \textbf{3.79} \\
        Original & 2.17 & \textbf{2.92} & 4.42 \\
        \hline
    \end{tabular}
    \label{tab:power_draw}
\end{table}

Lastly, we indicate the power draw of the full VIO system. In Fig. \ref{fig:power_draw} these three phases are depicted, the transition phases have been omitted for better readability.

The analysis in Table \ref{tab:power_draw} shows that turning on the optical flow unit leads to a 97\,mW higher power consumption when only running the camera capture. However, when subsequently turning on the VIO pipeline the extra power draw of the camera is offset by the reduced power draw of the full VIO pipeline, which consumes 630\,mW (14.24\%) less power than the original system on average.

\section{Discussion}
The conducted experiments show, that the VD56G3 sensor can be used in a VIO setup to reduce the processing load on the embedded processor on a small-scale mobile platform such as a UAV while maintaining a similar or higher accuracy as the original VINS-Mono pipeline. While we tested our hardware-software codesign approach in combination with VINS-Mono on a Raspberry Pi Compute Module 4, our method can be used in combination with other VIO pipelines and computing platforms to achieve similar results. Additionally, the approach could also be used in a pure visual odometry pipeline, either in a monocular or stereo camera setup. However, the use of an IMU is recommendable due to the complementary nature of cameras and IMUs \cite{he2020review}. The results shown in Table \ref{tab:rmse} further indicate, that the movement within a 4\,m by 4\,m square, which is given by the available motion capture system, is too small to leverage the potential of loop closure.

The observable differences between the tracking accuracy of both VINS-Mono implementations are partially caused by the parameterization of the optical flow unit. Furthermore, there are some architectural differences: The feature tracker of the original VINS-Mono pipeline enforces an even spread of feature points, whereas the focus in OF VINS-Mono is on the cornerness response of a feature and the spread of the features has a lower priority. Additionally, both pipelines maintain 150 feature tracks. When those features are lost, the original VINS-Mono implementation samples new features (i.e. feature tracks with length one). Instead of single features, OF VINS-Mono adds feature tracks of length two, already containing information regarding re-detectability thanks to the Hamming score, hence inherently applying outlier removal.

By analyzing different movement patterns individually, we verified that OF VINS-Mono can achieve consistent performance under varying conditions, when parameterized adequately (i.e. with the baseline parameter set and BRIEF 150). The ablation study furthermore showed that certain parameter sets can increase the performance under specific conditions, but do not work under other conditions (i.e. BRIEF 300), indicating that a dynamic parametrization based on the camera orientation or relative movement speed of the features might improve the system.

As the pre-processing operation of our proposed feature update step requires little processing time, the proposed solution has the potential to track high-speed motion, by operating the optical flow camera and the feature update at a high frame rate and keeping a moderate rate for the odometry estimation. This could allow leveraging the 200\,FPS at VGA and 300\,FPS at QVGA resolution of the optical flow camera in future work.

\section{Conclusion}
We presented the benefits of a high-speed low-power optical flow camera for embedded VIO applications where strict energy constraints and fast movements demand an energy-efficient and low-latency solution. To use the camera on low-power systems such as UAVs, a VIO pipeline based on VINS-Mono has been implemented on a resource-constrained processor such as the quad-core ARM Cortex-A72 processor. A new dataset has been collected with the optical flow sensor containing indoor recordings including ground truth poses. As the optical flow camera also returns the image data, analyses on full SLAM, including loop-closure have been conducted. For future work, we will evaluate the benefits of this camera and the low latency VIO in AR/VR and scale down the processing unit to target even lower-power embedded processors that can be embedded in nano-scale drones or miniaturized robots. 

\section*{Acknowledgment}
We would like to thank Sylvie Irigaray and Dominique Loyer at STMicroelectronics for the provided hardware and support during this project. We thank Paul Joseph for his preliminary results.

\addtolength{\textheight}{-6cm}

\bibliographystyle{IEEEtran}
\bibliography{references}

\begin{IEEEbiography}[{\includegraphics[width=1in,height=1.25in,clip,keepaspectratio]{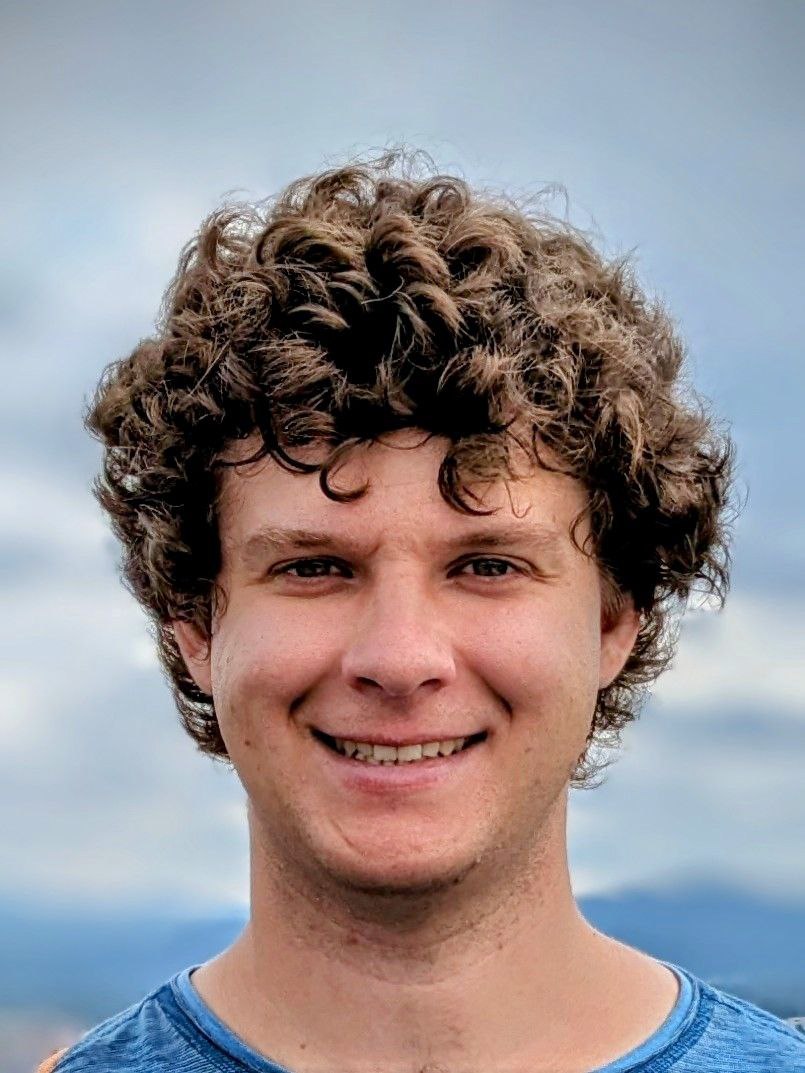}}]{Jonas Kühne}
(Graduate Student Member, IEEE) received the B.Sc. and M.Sc. degrees in electrical engineering and information technology from ETH Zürich, Zürich, Switzerland, in 2016 and 2018, respectively. Between 2019 and 2021, he worked for Agtatec AG which is part of Assa Abloy. \\
He is currently pursuing his Ph.D. degree with both the Integrated Systems Laboratory and the D-ITET Center for Project-Based Learning at ETH Zürich, Zürich, Switzerland. \\ 
His research interests include algorithm and hardware design for visual inertial odometry and SLAM on low-power embedded systems. 
\end{IEEEbiography}

\begin{IEEEbiography}[{\includegraphics[width=1in,height=1.25in,clip,keepaspectratio]{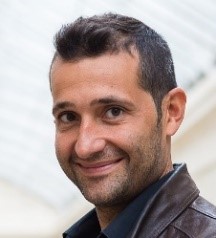}}]{Michele Magno}
(Senior Member, IEEE) received his master's and Ph.D. degrees in electronic engineering from the University of Bologna, Bologna, Italy, in 2004 and 2010, respectively. \\
Currently, he is a \emph{Privatdozent} at ETH Zurich, Zurich, Switzerland, where he is the Head of the Project-Based Learning Center. He has collaborated with several universities and research centers, such as Mid University Sweden, where he is a Guest Full Professor. He has published more than 150 articles in international journals and conferences, in which he got multiple best paper and best poster awards. The key topics of his research are wireless sensor networks, wearable devices, machine learning at the edge, energy harvesting, power management techniques, and extended lifetime of battery-operated devices.
\end{IEEEbiography}

\begin{IEEEbiography}[{\includegraphics[width=1in,height=1.25in,clip,keepaspectratio]{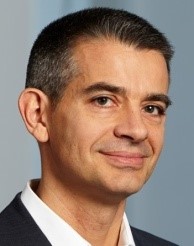}}]{Luca Benini}
(Fellow, IEEE) received the Ph.D. degree in electrical engineering from Stanford University, Stanford, CA, USA, in 1997.\\
He holds the Chair of Digital Circuits and Systems at ETH Zurich, Zurich, Switzerland, and is a Full Professor at the University of Bologna, Bologna, Italy. His current research interests include energy-efficient computing systems' design from embedded to high performance.\\
Dr. Benini is a fellow of the ACM and a member of the Academia Europaea. He was a recipient of the 2016 IEEE CAS Mac Van Valkenburg Award and the 2023 McCluskey  Award.
\end{IEEEbiography}

\end{document}